\documentclass[a4paper,fleqn]{cas-dc}

\usepackage[numbers]{natbib}
\usepackage{color}
\usepackage{graphicx}
\usepackage{subfigure}

\usepackage[T1]{fontenc}
\def\tsc#1{\csdef{#1}{\textsc{\lowercase{#1}}\xspace}}
\tsc{WGM}
\tsc{QE}

\begin{document}
\let\WriteBookmarks\relax
\def\floatpagepagefraction{1}
\def\textpagefraction{.001}
\let\printorcid\relax

\shorttitle{STGFormer: Spatio-Temporal GraphFormer for 3D Human Pose Estimation in Video}    
\shortauthors{Y.Liu et.al}  
\title [mode = title]{STGFormer: Spatio-Temporal GraphFormer for 3D Human Pose Estimation in Video} 



%

\author[1]{Yang Liu}[type=editor,
	style=english,
	auid=000]

\author[1]{Zhiyong Zhang}[type=editor,
	style=english,
	auid=000]

\cormark[1]


\ead{zhangzhy99@mail.sysu.edu.cn}



\affiliation[1]{organization={Sun Yat-sen University, School of Electronics and Communication Engineering},
            city={Shenzhen},
            state={Guangdong},
            country={China}}










\begin{abstract}
	The current methods of video-based 3D human pose estimation have achieved significant progress. However, they still face pressing challenges, such as the underutilization of spatiotemporal body-structure features in transformers and the inadequate granularity of spatiotemporal interaction modeling in graph convolutional networks, which leads to pervasive depth ambiguity in monocular 3D human pose estimation. To address these limitations, this paper presents the Spatio-Temporal GraphFormer framework (STGFormer) for 3D human pose estimation in videos. First, we introduce a Spatio-Temporal criss-cross Graph (STG) attention mechanism, designed to more effectively leverage the inherent graph priors of the human body within continuous sequence distributions while capturing spatiotemporal long-range dependencies. Next, we present a dual-path Modulated Hop-wise Regular GCN (MHR-GCN) to independently process temporal and spatial dimensions in parallel, preserving features rich in temporal dynamics and the original or high-dimensional representations of spatial structures. Furthermore, the module leverages modulation to optimize parameter efficiency and incorporates spatiotemporal hop-wise skip connections to capture higher-order information. Finally, we demonstrate that our method achieves state-of-the-art performance on the Human3.6M and MPI-INF-3DHP datasets.	
\end{abstract}




\maketitle

\section{Introduction}

3D human pose estimation predicts the coordinates of body keypoints in three dimensions from input images or videos. Compared to 2D pose estimation, it provides more comprehensive and accurate information about spatial relationships. In the field of computer vision, this technology has garnered widespread attention due to its potential applications in action recognition \cite{yin2024spatiotemporal, qiu2024multi}, human-computer interaction \cite{liu2022arhpe}, and autonomous driving \cite{wang2023learning, zheng2022multi}. Monocular 3D human pose estimation can primarily be divided into single-stage methods and two-stage methods. Single-stage methods directly estimate the end-to-end 3D human pose from inputs; two-stage methods initially detect the coordinates of keypoints and then regress to a 3D human pose based on the detected 2D keypoints. Currently, research indicates that the performance metrics of two-stage methods generally outperform those of single-stage methods, mainly due to the significant advancements in 2D pose estimation networks. For instance, ViTPose \cite{xu2022vitpose} utilizes a transformer as the backbone to extract features and employs a lightweight decoder for pose estimation. The proposed method has been demonstrated to be effective and flexible in 2D pose estimation. However, the second stage, which involves transitioning from 2D pose to 3D, remains unsatisfactory due to the ill-posed problem. This issue stems from the depth ambiguity inherent in estimating 3D coordinates from 2D, caused by a lack of depth prior information.

To address the aforementioned issues, a series of two-stage approaches for monocular 3D human pose estimation have been proposed that are based on backbones such as Convolutional Neural Networks (CNN) \cite{luvizon2022consensus, tang2023ftcm, zhou2021hemlets, wandt2022elepose}, Graph Convolutional Networks (GCN) \cite{yu2023gla, hua2022weakly, hassan2023regular, gong2023diffpose, wu2022hpgcn}, and transformers \cite{zhao2023poseformerv2, sun2023mixsynthformer, qian2023hstformer, einfalt2023uplift, zhang2022mixste, li2022mhformer, li2022exploiting}. For example, Einfalt et al. \cite{einfalt2023uplift} perform temporal upsampling within transformer blocks to process time-sparse 2D pose sequences, thereby reducing the model's size while still generating time-dense 3D pose estimates. StridedFormer \cite{li2022exploiting} reduces sequence redundancy by substituting the fully connected layers in the feedforward network of the Vanilla Transformer Encoder (VTE) with strided convolutions, thereby progressively decreasing the sequence length to lower computational costs. While transformers take advantage of the scalable capacity and high parallelism, offering promising improvements for pose estimation, they still face several challenges.

First, owing to the absence of structural priors in the transformer, its performance in 3D human pose estimation still trails behind the state-of-the-art methods based on GCNs. These graph-based methods effortlessly represent the human skeleton as a graph, where joints are regarded as nodes, and the bones between joints are considered as edges connecting these nodes. Currently, some improvements that integrate graphs with transformers have emerged \cite{zhao2021graformer, zhai2023hopfir, tang20233d}. For instance, STCFormer \cite{tang20233d} introduces a criss-cross attention strategy that, while mining the correlations across frame-level features, explores the relationships between joints across different frames. However, this method incorporates body structure graphs as inductive biases into the transformers. This approach affects the versatility of transformers, which are typically regarded as architectures with weak inductive biases, and has significant implications for pre-training and the scalability of large models.

To address the existing limitations, we propose a spatio-temporal criss-cross graph attention mechanism as part of our STGFormer architecture. This mechanism seamlessly integrates spatiotemporal graph structures into the attention layers, allowing the model to leverage prior knowledge about the human skeleton more effectively. By doing so, our approach enhances the ability to capture long-range dependencies across both spatial and temporal dimensions, ultimately improving performance in 3D human pose estimation.

Second, replacing Multi-Layer Perceptron (MLP) with GCN in transformer may enhance the ability to aggregate information throughout the entire graph structure. Integrating these two approaches allows the model to process graph data with greater precision and efficiency, thus optimizing performance from different aspects. However, in GCN computations, the computational cost grows exponentially with an increasing number of frames, rendering the approach impractical.

In this paper, we decouple temporal feature learning and spatial feature learning into a dual-path parallel operation within our MHR-GCN, which reduces the computational complexity from $\mathcal{O}\left(S \cdot T + (S-1)(T-1)\right)$ to $\mathcal{O}\left(S + (S-1)\right) + \mathcal{O}\left(T + (T-1)\right)$, where $S$ and $T$ denote the number of joints and frames, respectively. Moreover, the parallel dual-path module circumvents the sequential processing in serial architectures, which can lead to the premature compression of certain features. This approach effectively preserves both temporally dynamic features and the original or high-dimensional representations of spatial structures.

Third, in image-based 3D human pose estimation, mechanisms such as weight unsharing \cite{liu2020comprehensive} and weight modulation \cite{zou2021modulated} have been adopted to improve the efficiency of information exchange among nodes by preventing the use of uniformly shared transformation matrices. Meanwhile, higher-order dependencies have received increasing attention, with several studies addressing these dependencies by extending connections from first-order to higher-order neighbors \cite{xie2024hogformer, zou2020high}. Temporal higher-order dependencies originate from the system's dynamic characteristics, whereby the current state is influenced not only by the immediately preceding state but also by earlier states. However, the adoption of these methods in the temporal dimension has been neglected in video-based 3D human pose estimation.

To address this gap, we propose a novel spatiotemporal adjacency modulation mechanism within our MHR-GCN architecture, extending the original spatial modulation to the temporal dimension. This module further extends the original spatial modulation to the temporal dimension, preventing reliance on static or uniformly shared transformation matrices and thereby improving the flow of information across the network. In addition, we incorporate hop-wise spatiotemporal connections, which extend beyond immediate neighbors to capture higher-order dependencies across both space and time. By combining adjacency modulation with hop-wise connections, our MHR-GCN robustly models the dynamic characteristics of human motion and more effectively utilizes temporal higher-order dependencies in video-based 3D human pose estimation.

Our main contributions are summarized as follows:

(1) We propose a spatiotemporal criss-cross graph attention mechanism within our STGFormer architecture, integrating spatiotemporal graph structures into attention layers to more effectively leverage prior knowledge when capturing long-range dependencies.

(2) We introduce a dual-path modulated hop-wise regular GCN module and incorporate it into STGFormer, extracting first-order and higher-order spatiotemporal graph information from the features.

(3) We conduct extensive experiments on Human3.6M \cite{ionescu2013human3} and MPI-INF-3DHP \cite{mehta2017monocular} datasets for a comprehensive evaluation, showing that our proposed model performs well against the state-of-the-art methods.

\section{Related Work}

In recent years, both graph convolutional networks and transformers have made significant strides in 3D human pose estimation. As previously mentioned, we apply a two-stage approach to perform 3D human pose estimation from monocular image sequences or video inputs.

\subsection{GCN-based methods.}

The skeletal structure of the human body can be naturally represented as a graph. As a result, extensive research has been conducted on image-based 3D human pose estimation using GCNs, yielding significant results \cite{hassan2023regular, wu2022hpgcn, liu2020comprehensive, zou2021modulated, cai2019exploiting, zhao2019semantic}. To more effectively address the ill-posed problem of lifting 2D joint keypoints into 3D space, additional temporal information is extracted from sequential frames combined with spatial information related to body structure. For instance, the Global-Local Adaptive Graph Convolutional Network (GLA-GCN) \cite{yu2023gla} models the spatio-temporal structure using a graph representation and traces local joint features through independently connected layers. Zeng et al. \cite{zeng2021learning} developed a hop-aware hierarchical channel-squeezing fusion mechanism to efficiently extract relevant information from neighboring nodes in GCNs. Hu et al. \cite{hu2021conditional} represent the human skeleton as a directed graph to effectively reflect the hierarchical relationships among the nodes. Although graph-based methods effectively leverage the body’s structural information, they face a significant limitation in scalability. Adding more network layers does not improve model performance, and even commonly used residual connections have minimal impact.

\subsection{Transformer-based methods.}

In addition to GCN, transformers are also used to model spatio-temporal correlations for video-based 3D human pose estimation \cite{sun2023mixsynthformer, qian2023hstformer, einfalt2023uplift, zhang2022mixste, li2022mhformer, li2022exploiting, wang2019spatio, zheng20213d}. Specifically, MixSynthFormer \cite{sun2023mixsynthformer} combines synthesized spatial and temporal attention to effectively integrate inter-joint and inter-frame significance. HSTFormer \cite{qian2023hstformer} captures multi-level joint spatial-temporal correlations, ranging from local to global, through four transformer encoders and a fusion module. Li et al. \cite{li2022exploiting} utilized a vanilla transformer encoder to model long-range dependencies in 2D pose sequences, while a strided transformer encoder aggregates this long-range information into a single-vector representation, employing a hierarchical global and local approach. Transformers exhibit high scalability and are proficient at capturing long-range dependencies; however, they lack structural prior information. As a result, several methods have been developed that integrate graph and transformer. The Spatio-Temporal Criss-cross (STC) attention \cite{tang20233d} learns spatio-temporal features and incorporates the body's graph structure information as a biased input in the proposed framework. Zhai et al. \cite{zhai2023hopfir} proposed the Hop-wise GraphFormer, which enhances GCN-based pose estimation networks by grouping joints according to k-hop neighborhoods and capturing potential joint correlations across various joint synergies. Zhao et al. \cite{zhao2021graformer} designed a novel model, GraFormer, that combines graph convolution and transformers to exploit relationships among graph-structured 2D joints.

Our work also falls within the realm of 3D pose lifting, which employs a fusion of transformers and graphs. Unlike other approaches, our model directly integrates graph information into the attention layers, thereby significantly enhancing adaptability. Furthermore, it incorporates spatio-temporal hop-wise modulated GCNs within the transformer framework, effectively capturing high-order dependencies in both time and space. This approach complements the long-range dependencies typical of transformers, thus substantially enhancing the model's capacity to represent complex data structures.

\section{Method}
In this section, we first clarify the objectives of our learning task. Then, we present the main components of the proposed spatio-temporal graphformer for 3D human pose estimation, including the spatio-temporal criss-cross graph attention and the dual-path modulated hop-wise regular GCN module.

\subsection{Preliminary}

First, let us clarify the problem that needs to be solved in two-stage 3D human estimation based on deep learning. The training set for 3D human pose estimation contains paired data of $N$ joints $\mathcal{D} ={(\mathbf{x}_i,\mathbf{y}_i)},{i=1,\cdots, N}$, including 2D joint positions $\mathbf{X}=(\mathbf{x}_1,\cdots,\mathbf{x}_N)^{\mathsf{T}}\in\mathbb{R}^{N\times2}$ and their corresponding 3D joint positions $\mathbf{Y}=(\mathbf{y}_1,\cdots,\mathbf{y}_N)^{\mathsf{T}}\in\mathbb{R}^{N\times 3}$. The goal of human pose lifting is to minimize the loss function $l(f(\mathbf{x}_i),\mathbf{y}_i)$ of a regression model $f:\mathbf{X}\rightarrow\mathbf{Y}$ by learning the parameters.

Then, we provide some preliminary knowledge on human pose estimation using graph learning methods. Consider a graph $\mathcal{G}=(\mathcal{V},\mathcal{E})$, where $\mathcal{V}={1,\cdots,N}$ denotes the set of $N$ nodes and $\mathcal{E} \subseteq \mathcal{V} \times \mathcal{V}$ represents the set of edges. In the context of 3D human pose estimation, nodes correspond to human joints, and edges represent the connections between these joints. The adjacency matrix $\mathbf{A}=(a_{ij})$, with dimensions $N\times N$, contains elements $a_{ij}$ that represent the weights between node $i$ and $j$. These weights may be expressed as binary or real numbers. In our approach, we utilize a binary representation, where a value of $1$ indicates a connection $i\sim j$ between joints, while a value of $0$ signifies the absence of such a connection. The feature processing step transforms low-dimensional joints into high-dimensional features with a dimension of $F$. Each body can be represented by an $N \times F$ feature matrix $\mathbf{H}=(\mathbf{h}_1,\cdots,\mathbf{h}_N)^{\mathsf{T}}$, where each $\mathbf{h}_i$ is an $F$-dimensional feature vector corresponding to node $i$. For the inputs across consecutive frames $T$, they can be represented by a $T\times N\times F$ feature matrix $\mathbf{H}\in\mathbb{R}^{T\times N\times F}$.

Finally, it is essential to briefly introduce the transformer, which serves as the backbone of our framework. The transformer architecture consists of a set of transformer layers, each consisting of two components: a self-attention module and a position-wise feed-forward network. Let $\mathbf{H}\in \mathbb{R} ^{T\times N\times F}$ denote the input to the self-attention module where $F$ is the hidden dimension and $\mathbf{h}_{t,i}\in \mathbb{R} ^{1\times 1\times F}$ is the hidden representation at joint $i$ in the frame $t$. The input $\mathbf{H}$ is projected by three matrices $\mathbf{W}_Q$, $\mathbf{W}_K$ and $\mathbf{W}_V$ to the corresponding representations query $\mathbf{Q}$, key $\mathbf{K}$, and value $\mathbf{V}$. The self-attention is then computed as:
\begin{equation} 
	\begin{aligned}
		\mathbf{Q}&=\mathbf{HW}_Q, \mathbf{K}=\mathbf{HW}_K, \mathbf{V}=\mathbf{HW}_V  \\
		\mathbf{A}&=\frac{\mathbf{QK}^{\top}}{\sqrt{d_K}}, \mathrm{Attn}\left( \mathbf{H} \right) =\mathrm{softmax} \left( \mathbf{A} \right) \mathbf{V}
	\end{aligned}
	\label{eq8}
\end{equation}
where $\mathbf{A}$ is a matrix capturing the similarity between $\mathbf{Q}$ and $\mathbf{K}$, $\sqrt{d_K}$ represents the feature dimension of the query matrix.

\subsection{Overall Architecture}

\begin{figure*}[H]
	\centering
	\includegraphics[width=16cm]{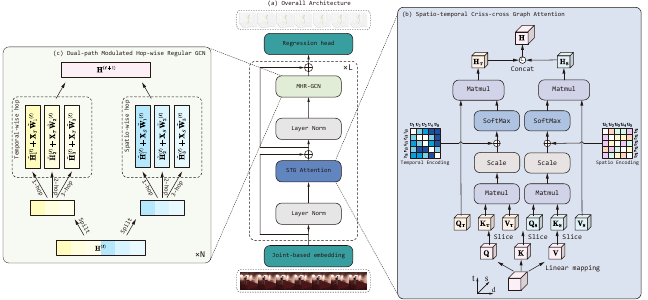}
	\caption{An overview of our proposed STGFormer. (a) It mainly consists of L sequential STGFormer blocks. (b) The architecture of our spatio-temporal criss-cross graph attention. (c) The architecture of our dual-path modulated hop-wise regular GCN module.}
	\vspace{-2mm}
	\label{fig1}
\end{figure*}

Figure \ref{fig1}(a) provides an overview of the proposed framework, which mainly comprises three parts: a joint-based embedding, spatio-temporal criss-cross graph transformer blocks, and a regression head. The joint-based embedding projects the input 2D coordinates of each joint into a higher-dimensional feature space. The STGFormer blocks integrate the STG attention and the dual-path MHR-GCN module to capture spatiotemporal and graph features, thereby updating the representation of each joint. The 3D coordinates are estimated from the learned features by a regression head.

\textbf{Joint-based embedding.}	

Given a 2D pose sequence represented as $\mathbf{P}_{2D}\in\mathbb{R}^{T\times N\times2}$, where $T$ denotes the number of frames and $N$ denotes the number of body joints per frame, we first map $\mathbf{P}_{2D}$ to high-dimensional embeddings using a joint-based embedding layer. This layer independently processes each 2D coordinate by applying a fully connected layer, followed by a GELU activation. As a result, the joint-based embedding layer produces features with dimensions $T\times N\times F$.

\textbf{Spatio-Temporal GraphFormer.}	 

The STGFormer block is derived from the transformer block in Eq.(\ref{eq8}), where the original multi-head self-attention layer is replaced with spatiotemporal criss-cross graph attention. Additionally, a novel dual-path MHR-GCN is incorporated into the STGFormer block to enhance its ability to model local structures. Detailed discussions on the STG attention and dual-path MHR-GCN are provided in Sections \ref{sec3.3} and \ref{sec3.4}, respectively.

\textbf{Regression head.}	

A linear regression head follows the STGFormer blocks to predict the 3D pose coordinates $\hat{\mathbf{P}}_{3D}\in\mathbb{R}^{T\times N\times 3}$. The entire architecture is optimized by minimizing the Mean Squared Error (MSE) between the predicted 3D coordinates $\hat{\mathbf{P}}_{3D}$ and the ground-truth 3D coordinates $\mathbf{P}_{3D}$:

\begin{equation} 
	\mathcal{L}=\left\|\hat{\mathbf{P}}_{3D}-\mathbf{P}_{3D}\right\|^{2}
	\label{eq9}
\end{equation} 

\subsection{Spatio-temporal Criss-cross Graph Attention} \label{sec3.3}	

Our attention mechanism is specifically designed to leverage prior knowledge of the human skeleton more effectively, ultimately improving performance in 3D human pose estimation. Inspired by \cite{tang20233d}, we capture spatial and temporal contexts in parallel across different channels, thereby avoiding the quadratic computational cost associated with full attention. Specifically, our mechanism encodes spatio-temporal graph structural relationships into the attention layer to capture structural information between nodes. While modeling the spatial and temporal relationships of the human body, a learnable embedding is assigned based on their structural relationships.	

For clarity, as shown in Figure \ref{fig1}(b), the input embedding $\mathbf{X}\in\mathbb{R}^{T\times N\times F}$ is first mapped to queries $\mathbf{Q}\in\mathbb{R}^{T\times N\times F}$, keys $\mathbf{K}\in\mathbb{R}^{T\times N\times F}$, and values $\mathbf{V}\in\mathbb{R}^{T\times N\times F}$, all sharing the same dimensions. These are subsequently split into two equal groups along the channel dimension. We refer to these split matrices as the time group $\{\mathbf{Q}_T,\mathbf{K}_T,\mathbf{V}_T\}$ and the space group $\{\mathbf{Q}_S,\mathbf{K}_S,\mathbf{V}_S\}$. Following this, separate self-attention modules are utilized to calculate the temporal and spatial correlations. Previous models required the explicit specification of distinct positions or the encoding of positional dependencies within layers. For sequential data, embeddings can be assigned to each position (i.e., absolute position encoding \cite{vaswani2017attention}) or encode the relative distance between any two positions in Transformer layers (i.e., relative position encoding \cite{raffel2020exploring}). However, for graphs, nodes are not sequentially arranged but are positioned in multidimensional spaces and connected by edges. 

To further integrate edge features into the attention layer and incorporate the structural information of graphs into the model, we propose a novel encoding technique in STGFormer. The temporal and spatial relationships of human joints are encoded in the softmax attention, which allows the model to accurately capture spatial dependencies in the graph.  Specifically, for each pair of nodes, we compute their connectivity with other nodes and then incorporate this information into the attention module. Equipped with these encodings, STGFormer is able to better model pairwise node relationships and represent the graph. This can be viewed as a refinement of the basic attention scores, endowing the model with the ability to distinguish different relative spatiotemporal positions and thereby better capture prior knowledge associated with spatial distance and temporal adjacency. For temporal encoding and spatial encoding, they are each respectively represented as follows: 
\begin{equation}
	A_{ij}^t=\frac{(h_i^tW_Q^t)(h_j^tW_K^t)^T}{\sqrt{d}}+{b^t}_{\psi(v_i,v_j)}
	\label{eq10}
\end{equation} 
\begin{equation}
	A_{ij}^s=\frac{(h_i^sW_Q^s)(h_j^sW_K^s)^T}{\sqrt{d}}+{b^s}_{\phi(u_i,u_j)}
	\label{eq11}
\end{equation} 
where $A_{ij}^t$ and $A_{ij}^s$ represent the attention scores between the $i$-th and $j$-th nodes in the temporal and spatial attention modules, respectively; $h_i^t$, $h_j^t$, $h_i^s$ and $h_j^s$ are the feature representation vectors of the $i$-th and $j$-th nodes corresponding to the temporal and spatial dimensions; $W_Q^t$, $W_K^t$, $W_Q^s$ and $W_K^s$ denote the learnable projection matrices that map the representations into the query and key spaces in the temporal and spatial attention branches; $\sqrt{d}$ represents the scaling factor; and ${b^t}_{\psi(v_i,v_j)}$ and ${b^s}_{\phi(u_i,u_j)}$ are learnable scalars, each indexed by its respective function $\psi(v_i,v_j)$ or $\phi(u_i,u_j)$.

Here we discuss the advantages of our proposed method. We introduce a function $b(\cdot)$ that quantifies the spatio-temporal relationships among elements within the graph, which is defined through the connectivity between nodes. Specifically, we depict the temporal and spatial connectivity relationships of the human body structure in continuous sequences through graphs, and encode these relationships as inputs for spatio-temporal attention on multiple channels in parallel. For example, if ${b}_{\psi(v_i,v_j)}$ is learned to be a decreasing function of $\psi(v_i,v_j)$, the model is more inclined to assign higher attention to nodes in close proximity and lower attention to those that are farther away.

Consequently, the two correlation modules described above operate in parallel and adhere to the self-attention scheme for feature contextualization. They calculate token-to-token affinities by contextualizing from specific axial perspectives, and effectively complement each other. Finally, we concatenate the outputs from both attention layers along the channel dimension:
\begin{equation}
	\begin{aligned}
		\mathbf{H}_T&=STGA_T(\mathbf{Q}_T,\mathbf{K}_T,\mathbf{V}_T) \\
		\mathbf{H}_S&=STGA_S(\mathbf{Q}_S,\mathbf{K}_S,\mathbf{V}_S) \\
		\mathbf{H}&=concat\left( \mathbf{H}_T,\mathbf{H}_S \right)
	\end{aligned}
	\label{eq12}
\end{equation} 
where $concat$ performs the concatenation. The resultant receptive field of our attention resembles a criss-cross of spatial and temporal axes, and stacking multiple STG attention blocks approximates complete spatio-temporal attention.

\subsection{Dual-path Modulated Hop-wise Regular GCN Module} \label{sec3.4}

To improve our understanding of spatiotemporal graph relationships by directly learning their interconnections, this paper introduces a novel GCN module. Our dual-path modulated hop-wise regular GCN module splits spatiotemporal graph convolution to independently learn features, thereby reducing computational complexity and preserving both temporally dynamic features and original (or high-dimensional) representations of spatial structures. The parallel dual-path strategy avoids the sequential processing typical of serial architectures, thereby mitigating the risk of prematurely compressing certain features. In addition, it employs a spatiotemporal hop-wise approach to capture higher-order dependencies and incorporates a residual connection to mitigate gradient vanishing and facilitate the iterative reuse of features.

Consider a graph convolutional layer in the vanilla GCN \cite{kipf2016semi} that transforms and aggregates input features according to the following equation:
\begin{equation}
	\mathbf{H}^{\prime}=\sigma \left(\mathbf{WH}\tilde{\mathbf{A}}\right)
	\label{eq1}
\end{equation}
where $\mathbf{H}^{\prime}$ is the updated feature matrix, $\sigma(\cdot)$ represents the activation function, $\mathbf{W}$ is the weight matrix, $\mathbf{H}$ is the input feature matrix, and $\tilde{\mathbf{A}}$ is the adjacency matrix. The $i$-th row of the output feature matrix can be expressed as:
\begin{equation}
	\mathbf{h}_{i}^{\prime}=\sigma \left(\sum_{j\in\mathcal{N}_i}{\mathbf{Wh}_j}\tilde{a}_{ij}\right)
	\label{eq2}
\end{equation}
where $\mathbf{h}_{i}^{\prime}$ is the updated feature vector of node $i$, $\mathbf{h}_j$ is the input feature vector of node $j$, $\mathcal{N}_i$ denotes the set of neighbors of node $j$, $\tilde{a}_{ij}$ is the element of the modified adjacency matrix corresponding to nodes $i$ and $j$.
However, the shared feature transformation weight in the vanilla GCN limits its ability to learn diverse relational patterns from different body joint nodes. Liu et al. \cite{liu2020comprehensive} address this limitation by utilizing a unique weight matrix $\mathbf{W}_j\in\mathbb{R}^{F^{\prime}\times F}$ for each node $j$. Nevertheless, as the number of joints and frames grows, this approach leads to a significant rise in the total number of parameters. To reduce model size, a weight modulation strategy \cite{zou2021modulated} is introduced, which applies a shared transformation but modulates it differently for each node, as expressed by the following equation:
\begin{equation}
	\mathbf{h}_{i}^{\prime} = \sigma \left(\sum_{j \in \mathcal{N}_i}{(}\mathbf{m}_j \odot \mathbf{W}) \mathbf{h}_j \tilde{a}_{ij}\right)
	\label{eq5}
\end{equation}
where $\mathbf{m}_j$ is a learnable modulation vector for each neighboring node $j$ and $\odot$ denotes element-wise multiplication.

\begin{figure*}[H]
	\centering
	\includegraphics[width=5.2cm]{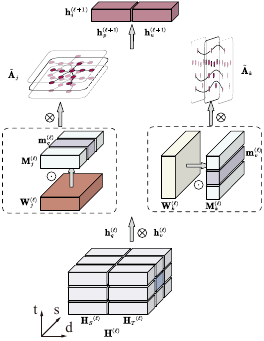}
	\caption{Illustration of the dual-path modulated processing in our MHR-GCN module. The symbol $\bigotimes$ denote matrix multiplication and $\bigodot$ denote element-wise multiplication, respectively.}
	\vspace{-0mm}
	\label{fig2}
\end{figure*}

As illustrated in Figure \ref{fig2}, we separate the temporal and spatial graph learning components into parallel pathways in our MHR-GCN. This decoupling mitigates the risk of feature entanglement and allows for independent optimization and regularization, leading to a more efficient and robust representation. Furthermore, by subsequently merging these two pathways, the model leverages complementary information that collectively enhances performance in tasks demanding detailed spatiotemporal understanding. In spatial modulation, a replica of the graph is created for each time step, while in temporal modulation, edges representing temporal relations are introduced to capture dependencies across time. As a result, by focusing on local spatial structures within individual frames and replicating the graph at every time step, the model can track variations in node connectivity and other spatial features with greater accuracy. The updated feature representations at layer $\ell$ for the spatial node $j$ and the temporal node $k$ can be expressed as follows:
\begin{equation}
	\begin{aligned}
		\tilde{\mathbf{H}}_{j}^{(\ell)}=&\sigma\left(\tilde{\mathbf{A}}_j(({\mathbf{H}_S^{(\ell)}}\mathbf{W}_{j}^{(\ell)})\odot \mathbf{M}_{j}^{(\ell )})\right)  \\
		\tilde{\mathbf{H}}_{k}^{(\ell)}=&\sigma\left(\tilde{\mathbf{A}}_k(({\mathbf{H}_T^{(\ell)}}\mathbf{W}_{k}^{(\ell)})\odot \mathbf{M}_{k}^{(\ell )})\right)
	\end{aligned}
	\label{eq13}
\end{equation}
where $\tilde{\mathbf{A}}_j$ and $\tilde{\mathbf{A}}_k$ are the spatial and temporal adjacency matrices, ${\mathbf{H}_S^{(\ell)}}$ and ${\mathbf{H}_T^{(\ell)}}$ are the input feature representations at layer $\ell$ for the spatial and temporal streams, respectively, and $\mathbf{W}_{j}^{(\ell )}$, $\mathbf{W}_{k}^{(\ell )}$ are the corresponding learnable weight matrices. The adjacency modulation matrices $\mathbf{M}_{k}^{(\ell )}$ and $\mathbf{M}_{j}^{(\ell )}$ allow the model presented in this paper to respond flexibly to data variations.

\begin{figure*}[H]
	\centering
	\vspace{-4mm}
	\includegraphics[width=12cm]{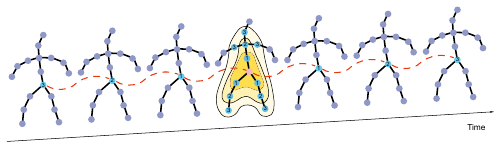}
	\vspace{-2mm}
	\caption{Illustration of the hop-wise concatenation in MHR-GCN for $K=3$ and $J=3$.}
	\vspace{-2mm}
	\label{fig3}
\end{figure*}

With increasing network depth, the model becomes more susceptible to suboptimal convergence (i.e., falling into local optima), and the resulting node representations tend to become overly smooth, making it difficult to distinguish features across different nodes. To address this limitation, we employ residual regular connections \cite{hassan2023regular} across layers to facilitate feature reuse in both the temporal and spatial paths. Moreover, as illustrated in Figure \ref{fig3}, the concept of hop-wise is introduced in MHR-GCN to capture high-order connection information and long-range dependencies. In this paper, we utilize $k$-hop temporal and $j$-hop spatial neighbors, as depicted in Figure \ref{fig1}(c), to further expand the receptive field.

Finally, the spatio-temporal layer-wise propagation rule of the regular dual-path hop-wise graph network with weight and adjacency modulation is formulated as follows, where $\mathbf{H}_{S}^{(\ell)}\in\mathbb{R}^{N\times F}$ and $\mathbf{H}_{T}^{(\ell)}\in\mathbb{R}^{N\times F}$ represent as the input feature matrices at the $\ell$-th layer: 
\begin{equation}
	\begin{aligned}
		\mathbf{H}_{S}^{(\ell +1)}=&\sigma \left( \underset{j=1}{\overset{J}{\boxplus}}\left( \tilde{\mathbf{H}}_{j}^{(\ell )}+\mathbf{X}_S{\tilde{\mathbf{W}}_j}^{(\ell )} \right) \right)  \\
		\mathbf{H}_{T}^{(\ell +1)}=&\sigma \left( \underset{k=1}{\overset{K}{\boxplus}}\left( \tilde{\mathbf{H}}_{k}^{(\ell )}+\mathbf{X}_T{\tilde{\mathbf{W}}_k}^{(\ell )} \right) \right)
	\end{aligned}	
	\label{eq14}
\end{equation} 
where $J$ and $K$ represent the hop range considered in the spatial and temporal domain, $\tilde{\mathbf{H}}_{j}^{(\ell)}$ and $\tilde{\mathbf{H}}_{k}^{(\ell)}$ are the intermediate representations obtained at layer $\ell$ through $j$-hop or $k$-hop neighbor aggregation, $\mathbf{X}_S{\tilde{\mathbf{W}}_j}^{(\ell )}$ and $\mathbf{X}_T{\tilde{\mathbf{W}}_k}^{(\ell )}$ are the regular connection, $\mathbf{H}_{S}^{(\ell+1)}\in\mathbb{R}^{N\times F}$, $\mathbf{H}_{T}^{(\ell+1)}\in\mathbb{R}^{N\times F}$ are the output feature matrix, $\boxplus$ denotes concatenation. The output feature vectors $\mathbf{h}_{p}^{(\ell +1)}$ and $\mathbf{h}_{u}^{(\ell +1)}$ of node $p$, $u$ on the spatial and temporal paths can be represented as follows: 
\begin{equation}
	\begin{aligned}
		\mathbf{h}_{p}^{(\ell +1)}=&\sigma \left( \sum_{q\in \mathcal{N} _p}{\tilde{a}_{pq}}\mathbf{h}_{q}^{(\ell )}({\mathbf{W}_j}^{(\ell )}\odot \mathbf{m}_{q}^{(\ell )})+\mathbf{x}_p\widetilde{\mathbf{W}_j}^{(\ell )} \right)  \\
		\mathbf{h}_{u}^{(\ell +1)}=&\sigma \left( \sum_{v\in \mathcal{N} _u}{\tilde{a}_{uv}}\mathbf{h}_{v}^{(\ell )}({\mathbf{W}_k}^{(\ell )}\odot \mathbf{m}_{v}^{(\ell )})+\mathbf{x}_u\widetilde{\mathbf{W}_k}^{(\ell )} \right)
	\end{aligned}	
	\label{eq15}
\end{equation} 
where $\mathcal{N}_p$ and $\mathcal{N}_u$ respectively denote the neighbor sets of node $p$ and node $u$, $\tilde{a}_{pq}$ and $\tilde{a}_{uv}$ represent the adjacency coefficients between the node pairs $(p, q)$ and $(u, v)$, $\mathbf{m}_{q}$ and $\mathbf{m}_{v}$ are the modulation vectors of node $q$ and node $v$, $\mathbf{x}_p$ and $\mathbf{x}_u$ respectively represent the initial input features of node $p$ and node $u$. Thus, the equation of the $\ell +1$ layer feature is reformulated as
\begin{equation}
	\mathbf{H}^{(\ell +1)}=\sigma \left( \underset{j=1}{\overset{J}{\boxplus}}\left( \tilde{\mathbf{H}}_{j}^{(\ell )}+\mathbf{X}_S{\tilde{\mathbf{W}}_j}^{(\ell )} \right) \right) \boxplus \sigma \left( \underset{k=1}{\overset{K}{\boxplus}}\left( \tilde{\mathbf{H}}_{k}^{(\ell )}+\mathbf{X}_T{\tilde{\mathbf{W}}_k}^{(\ell )} \right) \right)	
	\label{eq16}
\end{equation}

\section{Experiments}

In this section, we first introduce the benchmark datasets, Human3.6M \cite{ionescu2013human3} and MPI-INF-3DHP \cite{mehta2017monocular}, and the evaluation metrics used to assess the proposed framework, along with the experimental setup. We then evaluate the framework and compare it with current state-of-the-art methods, presenting both quantitative and qualitative results. Finally, we conduct ablation studies to assess the impact of each proposed design.

\subsection{Experimental Setup}

\noindent\textit{1) Datasets and Evaluation Metrics:}

\textbf{Human3.6M} \cite{ionescu2013human3} is a widely used dataset for 3D human pose estimation, comprising 3.6 million poses captured using RGB and ToF cameras from multiple viewpoints in real-world settings. Following the standard protocol, we use five subjects (S1, S5, S6, S7, S8) for training and two subjects (S9, S11) for evaluation. We evaluate performance using Mean Per Joint Position Error (MPJPE) under two protocols: Protocol 1 (P1), which aligns the root joints before calculating MPJPE, and Protocol 2 (P2), which involves Procrustes-MPJPE (PA-MPJPE), where the ground truth and the predicted pose are aligned through a rigid transformation.

\textbf{MPI-INF-3DHP} \cite{mehta2017monocular} is a large-scale dataset with over 2K videos featuring joint annotations of 13 keypoints in outdoor scenes, designed for both 2D and 3D human pose estimation. The ground truth is obtained through a multi-camera setup and a marker-less MoCap system. The evaluation metrics include MPJPE, Percentage of Correct Keypoints (PCK) at a threshold of 150 mm, and Area Under the Curve (AUC).

\noindent\textit{2) Baseline Methods:}

We evaluate our model by comparing it with state-of-the-art video-based 3D human pose estimation methods, including transformer-based approaches such as StridedFormer \cite{li2022exploiting}, MHFormer \cite{li2022mhformer}, MixSTE \cite{zhang2022mixste}, P-STMO \cite{shan2022p}, HSTFormer \cite{qian2023hstformer}, Einfalt et al. \cite{einfalt2023uplift}, and PoseFormerV2 \cite{zhao2023poseformerv2}, as well as GCN-based methods like UGCN \cite{wang2020motion} and GLA-GCN \cite{yu2023gla}. We also compare our model with hybrid methods combining graphs and transformers, such as STCFormer \cite{tang20233d}, Motionbert \cite{zhu2023motionbert}, and MotionAGFormer \cite{mehraban2024motionagformer}.

\noindent\textit{3) Implementation Details:}

For the experiments, we use three types of 2D pose sequences as input: estimated 2D poses from the pre-trained CPN \cite{chen2018cascaded}, Stacked Hourglass \cite{xu2021graph}, and real 2D poses (ground truth). The model is trained for 40 epochs using the Adam optimizer with a batch size of 128. The learning rate starts at 0.001 and decays by 0.97 per epoch. Our architecture is configured with L = 6 transformer layers, F = 256 feature embedding dimensions, H = 8 self-attention heads, and N = 2 MHR-GCN layers. The model is implemented using PyTorch (version 2.1.0) and runs on an RTX 4090 GPU with driver version 525.105.17 and CUDA version 12.0.

\subsection{Results and Analysis}

\noindent\textit{1) Quantitative Results:}

\textbf{Performance Comparison on Human3.6M.}
We conducted a comparative analysis of our method against state-of-the-art approaches on the Human3.6M dataset. Tables \ref{tab:table1} and \ref{tab:table2} present our performance in terms of P1 and P2 errors, respectively, using CPN and SH pre-trained 2D coordinates as inputs. Our method establishes a new state-of-the-art for a 243-frame setting, outperforming transformer-based, GCN-based, and graph-augmented transformer methods. For CPN, benefiting from our model's strong ability to learn body information, our approach shows improvements over transformer-based methods such as StridedFormer \cite{li2022exploiting}, MHFormer \cite{li2022mhformer}, and MixSTE \cite{zhang2022mixste}, with gains of 3.4 mm, 2.7 mm, and 0.6 mm in P1 error, and 3.5 mm, 2.7 mm, and 0.9 mm in P2 error, respectively. Compared to UGCN \cite{wang2020motion}, GLA-GCN \cite{yu2023gla}, and STCFormer \cite{tang20233d}, our method shows enhancements of 5.3 mm, 4.1 mm, and 0.7 mm in P1 error. Notably, in a series of actions, our method achieves the best performance recorded to date in 10 categories. When comparing with Motionbert \cite{zhu2023motionbert} and MotionAGFormer \cite{mehraban2024motionagformer} using the more powerful 2D detector Stacked Hourglass (SH) \cite{xu2021graph}, our method further improves by approximately 2.3 mm (5.7\%) in P1 error and 0.9 mm (2.8\%) in P2 error. Our model significantly outperforms all other methods, achieving an approximate 0.4 mm improvement in average MPJPE compared to MotionAGFormer \cite{mehraban2024motionagformer}. Table \ref{tab:table3} shows the performance of our method on P1 error when using ground-truth 2D poses as input. Compared to CPN or SH 2D poses, the use of ground-truth 2D poses eliminates noise interference, enabling each model to reach peak performance. The trend of P1 error remains consistent, with our method leading all competitors in the 243-frame input case, achieving a P1 error of 21.4 mm, which outperforms the current leading model, MixSTE \cite{zhang2022mixste}, by 0.2 mm.

\textbf{Performance Comparison on MPI-INF-3DHP.}
To assess the generalizability of our proposed framework, we tested it on the MPI-INF-3DHP dataset, which is known for its challenging backgrounds. Given the shorter video sequences, we configured the number of input frames to be 9, 27, and 81, respectively. Detailed performance comparisons are provided in Table \ref{tab:table4}. As observed in the results on Human3.6M, the optimal performance is achieved with 81 frames, reaching a PCK of 98.8, AUC of 84.1, and P1 error of 19.2 mm. This performance outperforms the current state-of-the-art model by 0.1 in PCK and 0.2 in AUC. These results underscore the effectiveness of the proposed method in handling more complex scenarios. In our experiments, the MPI-INF-3DHP dataset primarily consists of complex outdoor scenes, with a focus on evaluating the accuracy of algorithm results. The experimental findings show that the accuracy metrics, AUC and PCK, outperform SOTA methods, including MotionAGFormer. On the other hand, for the precision metric MPJPE, comparisons on the Human3.6M dataset are more extensive, and our method’s MPJPE and P-MPJPE metrics outperform MotionAGFormer.

\begin{table*}[H]
	\centering
	\caption{Performance Comparison on Human3.6M utilizes 2D poses estimated by CPN and SH as inputs under Protocol 1. Methods marked with * denote transformer-based approaches, \textasciicircum indicates GCN-based methods, and $\dagger$ represents graph-augmented transformer methods.}
	\vspace{0mm}
	\resizebox{0.95\textwidth}{!}{
		\begin{tabular}{cccccccccccccccccc}
			\hline \hline
			Method                                  & Publication & Dir. & Disc. & Eat. & Greet & Phone & Photo & Pose & Purch. & Sit  & SitD. & Smoke & Wait & WalkD. & Walk & WlkT. & Avg. \\ \hline
			StridedFormer \cite{li2022exploiting}* (CPN, T=243)          & TMM’22      & 40.3 & 43.3  & 40.2 & 42.3  & 45.6  & 52.3  & 41.8 & 40.5   & 55.9 & 60.6  & 44.2  & 43.0 & 44.2   & 30.0 & 30.2  & 43.7 \\
			MHFormer \cite{li2022mhformer}* (CPN, T=351)               & CVPR’22     & 39.2 & 43.1  & 40.1 & 40.9  & 44.9  & 51.2  & 40.6 & 41.3   & 53.5 & 60.3  & 43.7  & 41.1 & 43.8   & 29.8 & 30.6  & 43.0 \\
			MixSTE \cite{zhang2022mixste}* (CPN, T=243)                 & CVPR’22     & \textcolor{red}{37.6} & \textcolor{blue}{40.9}  & 37.3 & 39.7  & \textcolor{red}{42.3}  & \textcolor{blue}{49.9}  & 40.1 & 39.8   & 51.7 & \textcolor{red}{55.0}  & 42.1  & 39.8 & 41.0   & 27.9 & \textcolor{red}{27.9}  & \textcolor{blue}{40.9} \\
			P-STMO \cite{shan2022p}* (CPN, T=243)                 & ECCV’22     & 38.9 & 42.7  & 40.4 & 41.1  & 45.6  & \textcolor{red}{49.7}  & 40.9 & 39.9   & 55.5 & 59.4  & 44.9  & 42.2 & 42.7   & 29.4 & 29.4  & 42.8 \\
			HSTFormer \cite{qian2023hstformer}* (CPN, T=81)               & arXiv’23    & 39.5 & 42.0  & 39.9 & 40.8  & 44.4  & 50.9  & 40.9 & 41.3   & 54.7 & 58.8  & 43.6  & 40.7 & 43.4   & 30.1 & 30.4  & 42.7 \\
			Einfalt et al. \cite{einfalt2023uplift}* (CPN, T=351)         & WACV’23     & 39.6 & 43.8  & 40.2 & 42.4  & 46.5  & 53.9  & 42.3 & 42.5   & 55.7 & 62.3  & 45.1  & 43.0 & 44.7   & 30.1 & 30.8  & 44.2 \\
			UGCN \cite{wang2020motion}\textasciicircum (CPN, T=96)     & ECCV’20     & 41.3 & 43.9  & 44.0 & 42.2  & 48.0  & 57.1  & 42.2 & 43.2   & 57.3 & 61.3  & 47.0  & 43.5 & 47.0   & 32.6 & 31.8  & 45.6 \\
			GLA-GCN \cite{yu2023gla}\textasciicircum (CPN, T=243) & CVPR’23     & 41.3 & 44.3  & 40.8 & 41.8  & 45.9  & 54.1  & 42.1 & 41.5   & 57.8 & 62.9  & 45.0  & 42.8 & 45.9   & 29.4 & 29.9  & 44.4 \\
			STCFormer \cite{tang20233d}$\dagger$ (CPN, T=243)             & CVPR’23     & 39.6 & 41.6  & 37.4 & \textcolor{blue}{38.8}  & 43.1  & 51.1  & \textcolor{blue}{39.1} & \textcolor{blue}{39.7}   & 51.4 & 57.4  & \textcolor{blue}{41.8}  & \textcolor{blue}{38.5} & 40.7   & \textcolor{blue}{27.1} & 28.6  & 41.0 \\
			\cellcolor[rgb]{.851,.851,.851}STGFormer (Ours, CPN, T=81)                            & \cellcolor[rgb]{.851,.851,.851}            & \cellcolor[rgb]{.851,.851,.851}39.5 & \cellcolor[rgb]{.851,.851,.851}41.8  & \cellcolor[rgb]{.851,.851,.851}\textcolor{blue}{37.2} & \cellcolor[rgb]{.851,.851,.851}39.2  & \cellcolor[rgb]{.851,.851,.851}43.2  & \cellcolor[rgb]{.851,.851,.851}50.5  & \cellcolor[rgb]{.851,.851,.851}40.3 & \cellcolor[rgb]{.851,.851,.851}40.2   & \cellcolor[rgb]{.851,.851,.851}\textcolor{blue}{51.1} & \cellcolor[rgb]{.851,.851,.851}56.6  & \cellcolor[rgb]{.851,.851,.851}42.3  & \cellcolor[rgb]{.851,.851,.851}39.0 & \cellcolor[rgb]{.851,.851,.851}\textcolor{blue}{40.3}   & \cellcolor[rgb]{.851,.851,.851}27.7 & \cellcolor[rgb]{.851,.851,.851}\textcolor{blue}{28.1}  & \cellcolor[rgb]{.851,.851,.851}41.1 \\
			\cellcolor[rgb]{.851,.851,.851}STGFormer (Ours, CPN, T=243)                           & \cellcolor[rgb]{.851,.851,.851}            & \cellcolor[rgb]{.851,.851,.851}\textcolor{blue}{38.4} & \cellcolor[rgb]{.851,.851,.851}\textcolor{red}{40.7}  & \cellcolor[rgb]{.851,.851,.851}\textcolor{red}{37.1} & \cellcolor[rgb]{.851,.851,.851}\textcolor{red}{38.1}  & \cellcolor[rgb]{.851,.851,.851}\textcolor{blue}{42.7}  & \cellcolor[rgb]{.851,.851,.851}\textcolor{blue}{49.9}  & \cellcolor[rgb]{.851,.851,.851}\textcolor{red}{39.0} & \cellcolor[rgb]{.851,.851,.851}\textcolor{red}{39.3}   & \cellcolor[rgb]{.851,.851,.851}\textcolor{red}{50.4} & \cellcolor[rgb]{.851,.851,.851}\textcolor{blue}{56.0}  & \cellcolor[rgb]{.851,.851,.851}\textcolor{red}{41.4}  & \cellcolor[rgb]{.851,.851,.851}\textcolor{red}{38.3} & \cellcolor[rgb]{.851,.851,.851}\textcolor{red}{39.1}   & \cellcolor[rgb]{.851,.851,.851}\textcolor{red}{26.9} & \cellcolor[rgb]{.851,.851,.851}\textcolor{blue}{28.1}  & \cellcolor[rgb]{.851,.851,.851}\textcolor{red}{40.3} \\ \hline			
			Motionbert \cite{zhu2023motionbert}$\dagger$ (SH, T=243)             & ICCV'23     & \textcolor{blue}{36.3} & 38.7  & 38.6 & \textcolor{blue}{33.6}  & 42.1  & 50.1 & \textcolor{blue}{36.2} & 35.7   & \textcolor{blue}{50.1} & 56.6  & 41.3  & 37.4 & 37.7   & \textcolor{blue}{25.6} & 26.5  & 39.2 \\
			MotionAGFormer \cite{mehraban2024motionagformer}$\dagger$ (SH, T=243)             & WACV’24     & 36.4 & \textcolor{blue}{38.4}  & 36.8 & \textcolor{red}{32.9}  & \textcolor{blue}{40.9}  & \textcolor{blue}{48.5}  & 36.6 & \textcolor{blue}{34.6}   & 51.7 & \textcolor{blue}{52.8}  & \textcolor{blue}{41.0}  & \textcolor{red}{36.4} & \textcolor{red}{36.5}   & 26.7 & 27.0  & \textcolor{blue}{38.4} \\
			\cellcolor[rgb]{.851,.851,.851}STGFormer (Ours, SH, T=81)                           & \cellcolor[rgb]{.851,.851,.851}            & \cellcolor[rgb]{.851,.851,.851}38.5 & \cellcolor[rgb]{.851,.851,.851}39.1  & \cellcolor[rgb]{.851,.851,.851}\textcolor{blue}{36.2} & \cellcolor[rgb]{.851,.851,.851}37.8  & \cellcolor[rgb]{.851,.851,.851}42.1  & \cellcolor[rgb]{.851,.851,.851}49.9  & \cellcolor[rgb]{.851,.851,.851}36.4 & \cellcolor[rgb]{.851,.851,.851}35.7   & \cellcolor[rgb]{.851,.851,.851}50.8 & \cellcolor[rgb]{.851,.851,.851}54.5  & \cellcolor[rgb]{.851,.851,.851}41.1  & \cellcolor[rgb]{.851,.851,.851}37.1 & \cellcolor[rgb]{.851,.851,.851}38.0   & \cellcolor[rgb]{.851,.851,.851}26.8 & \cellcolor[rgb]{.851,.851,.851}\textcolor{blue}{26.1}  & \cellcolor[rgb]{.851,.851,.851}39.3 \\
			\cellcolor[rgb]{.851,.851,.851}STGFormer (Ours, SH, T=243)                          & \cellcolor[rgb]{.851,.851,.851}            & \cellcolor[rgb]{.851,.851,.851}\textcolor{red}{35.9} & \cellcolor[rgb]{.851,.851,.851}\textcolor{red}{37.6}  & \cellcolor[rgb]{.851,.851,.851}\textcolor{red}{35.4} & \cellcolor[rgb]{.851,.851,.851}36.6  & \cellcolor[rgb]{.851,.851,.851}\textcolor{red}{40.5}  & \cellcolor[rgb]{.851,.851,.851}\textcolor{red}{48.0}  & \cellcolor[rgb]{.851,.851,.851}\textcolor{red}{35.7} & \cellcolor[rgb]{.851,.851,.851}\textcolor{red}{34.2}   & \cellcolor[rgb]{.851,.851,.851}\textcolor{red}{49.7} & \cellcolor[rgb]{.851,.851,.851}\textcolor{red}{52.4}  & \cellcolor[rgb]{.851,.851,.851}\textcolor{red}{39.6}  & \cellcolor[rgb]{.851,.851,.851}\textcolor{blue}{36.8} & \cellcolor[rgb]{.851,.851,.851}\textcolor{blue}{37.2}   & \cellcolor[rgb]{.851,.851,.851}\textcolor{red}{24.9} & \cellcolor[rgb]{.851,.851,.851}\textcolor{red}{25.7}  & \cellcolor[rgb]{.851,.851,.851}\textcolor{red}{38.0} \\ \hline \hline
		\end{tabular}
	}
	\label{tab:table1}
	\vspace{0mm}
\end{table*}

\begin{table*}[H]
	\centering
	\caption{Performance Comparison on Human3.6M utilizes 2D poses estimated by CPN and SH as inputs under Protocol 2. Methods marked with * denote transformer-based approaches, \textasciicircum indicates GCN-based methods, and $\dagger$ represents graph-augmented transformer methods.}
	\vspace{0mm}
	\resizebox{0.95\textwidth}{!}{
		\begin{tabular}{cccccccccccccccccc}
			\hline \hline
			Method                                  & Publication & Dir. & Disc. & Eat. & Greet & Phone & Photo & Pose & Purch. & Sit  & SitD. & Smoke & Wait & WalkD. & Walk & WlkT. & Avg. \\ \hline
			StridedFormer \cite{li2022exploiting}* (CPN, T =243)         & TMM’22      & 32.7 & 35.5  & 32.5 & 35.4  & 35.9  & 41.6  & 33.0 & 31.9   & 45.1 & 50.1  & 36.3  & 33.5 & 35.1   & 23.9 & 25.0  & 35.2 \\
			MHFormer \cite{li2022mhformer}* (CPN, T=351)               & CVPR’22     & 31.5 & 34.9  & 32.8 & 33.6  & 35.3  & 39.6  & 32.0 & 32.2   & 43.5 & 48.7  & 36.4  & 32.6 & 34.3   & 23.9 & 25.1  & 34.4 \\
			MixSTE \cite{zhang2022mixste}* (CPN, T=243)                 & CVPR’22     & 30.8 & \textcolor{blue}{33.1}  & 30.3 & 31.8  & 33.1  & 39.1  & 31.1 & 30.5   & 42.5 & \textcolor{blue}{44.5}  & 34.0  & 30.8 & 32.7   & 22.1 & 22.9  & 32.6 \\
			P-STMO \cite{shan2022p}* (CPN, T=243)                 & ECCV’22     & 31.3 & 35.2  & 32.9 & 33.9  & 35.4  & 39.3  & 32.5 & 31.5   & 44.6 & 48.2  & 36.3  & 32.9 & 34.4   & 23.8 & 23.9  & 34.4 \\
			HSTFormer \cite{qian2023hstformer}* (CPN, T=81)               & arXiv’23    & 31.1 & 33.7  & 33.0 & 33.2  & 33.6  & 38.8  & 31.9 & 31.5   & 43.7 & 46.3  & 35.7  & 31.5 & 33.1   & 24.2 & 24.5  & 33.7 \\
			Einfalt et al. \cite{einfalt2023uplift}* (CPN, T=351)         & WACV’23     & 32.7 & 36.1  & 33.4 & 36.0  & 36.1  & 42.0  & 33.3 & 33.1   & 45.4 & 50.7  & 37.0  & 34.1 & 35.9   & 24.4 & 25.4  & 35.7 \\
			UGCN \cite{wang2020motion}\textasciicircum (CPN, T=96)     & ECCV’20     & 32.9 & 35.2  & 35.6 & 34.4  & 36.4  & 42.7  & 31.2 & 32.5   & 45.6 & 50.2  & 37.3  & 32.8 & 36.3   & 26.0 & 23.9  & 35.5 \\
			GLA-GCN \cite{yu2023gla}\textasciicircum (CPN, T=243) & CVPR’23     & 32.4 & 35.3  & 32.6 & 34.2  & 35.0  & 42.1  & 32.1 & 31.9   & 45.5 & 49.5  & 36.1  & 32.4 & 35.6   & 23.5 & 24.7  & 34.8 \\
			STCFormer \cite{tang20233d}$\dagger$ (CPN, T=243)             & CVPR’23     & \textcolor{blue}{29.5} & 33.2  & 30.6 & \textcolor{blue}{31.0}  & \textcolor{blue}{33.0} & 38.0  & 30.4 & \textcolor{red}{29.4}   & 41.8 & 45.2  & 33.6  & 29.5 & \textcolor{blue}{31.6}   & \textcolor{red}{21.3} & \textcolor{red}{22.6}  & \textcolor{blue}{32.0} \\
			\cellcolor[rgb]{.851,.851,.851}STGFormer (Ours, CPN, T=81)                            & \cellcolor[rgb]{.851,.851,.851}            & \cellcolor[rgb]{.851,.851,.851}29.6 & \cellcolor[rgb]{.851,.851,.851}33.2  & \cellcolor[rgb]{.851,.851,.851}\textcolor{blue}{30.2} & \cellcolor[rgb]{.851,.851,.851}31.3  & \cellcolor[rgb]{.851,.851,.851}\textcolor{blue}{33.0}  & \cellcolor[rgb]{.851,.851,.851}\textcolor{blue}{37.7}  & \cellcolor[rgb]{.851,.851,.851}\textcolor{red}{30.1} & \cellcolor[rgb]{.851,.851,.851}30.4   & \cellcolor[rgb]{.851,.851,.851}\textcolor{blue}{41.7} & \cellcolor[rgb]{.851,.851,.851}44.6  & \cellcolor[rgb]{.851,.851,.851}\textcolor{blue}{33.5}  & \cellcolor[rgb]{.851,.851,.851}\textcolor{blue}{29.5} & \cellcolor[rgb]{.851,.851,.851}31.7   & \cellcolor[rgb]{.851,.851,.851}22.0 & \cellcolor[rgb]{.851,.851,.851}23.1  & \cellcolor[rgb]{.851,.851,.851}32.1 \\
			\cellcolor[rgb]{.851,.851,.851}STGFormer (Ours, CPN, T=243)                           & \cellcolor[rgb]{.851,.851,.851}            & \cellcolor[rgb]{.851,.851,.851}\textcolor{red}{29.3} & \cellcolor[rgb]{.851,.851,.851}\textcolor{red}{33.0}  & \cellcolor[rgb]{.851,.851,.851}\textcolor{red}{29.9} & \cellcolor[rgb]{.851,.851,.851}\textcolor{red}{30.6}  & \cellcolor[rgb]{.851,.851,.851}\textcolor{red}{32.5}  & \cellcolor[rgb]{.851,.851,.851}\textcolor{red}{37.5}  & \cellcolor[rgb]{.851,.851,.851}\textcolor{blue}{30.3} & \cellcolor[rgb]{.851,.851,.851}\textcolor{blue}{29.6}   & \cellcolor[rgb]{.851,.851,.851}\textcolor{red}{41.0} & \cellcolor[rgb]{.851,.851,.851}\textcolor{red}{44.1}  & \cellcolor[rgb]{.851,.851,.851}\textcolor{red}{33.2}  & \cellcolor[rgb]{.851,.851,.851}\textcolor{red}{29.1} & \cellcolor[rgb]{.851,.851,.851}\textcolor{red}{30.9}   & \cellcolor[rgb]{.851,.851,.851}\textcolor{blue}{21.7} & \cellcolor[rgb]{.851,.851,.851}\textcolor{blue}{22.8}  & \cellcolor[rgb]{.851,.851,.851}\textcolor{red}{31.7} \\ \hline
			Motionbert \cite{zhu2023motionbert}$\dagger$ (SH, T=243)            & ICCV'23     & 30.8 & 32.8  & 32.4 & 28.7  & 34.3  & 38.9 & 30.1 & 30.0   & 42.5 & 49.7  & 36.0  & 30.8 & \textcolor{red}{22.0}   & 31.7 & \textcolor{blue}{23.0}  & 32.9 \\
			MotionAGFormer \cite{mehraban2024motionagformer}$\dagger$ (SH, T=243)             & WACV’24     & 30.6 & \textcolor{blue}{32.6}  & 32.2 & \textcolor{blue}{28.2}  & 33.8  & 38.6 & 30.5 & \textcolor{blue}{29.9}   & 43.3 & 47.0  & 35.2  & \textcolor{blue}{29.8} & 31.4   & 22.7 & 23.5  & 32.6 \\
			\cellcolor[rgb]{.851,.851,.851}STGFormer (Ours, SH, T=81)                            & \cellcolor[rgb]{.851,.851,.851}            & \cellcolor[rgb]{.851,.851,.851}\textcolor{blue}{29.4} & \cellcolor[rgb]{.851,.851,.851}33.1  & \cellcolor[rgb]{.851,.851,.851}\textcolor{blue}{29.3} & \cellcolor[rgb]{.851,.851,.851}28.9  & \cellcolor[rgb]{.851,.851,.851}\textcolor{blue}{32.4}  & \cellcolor[rgb]{.851,.851,.851}\textcolor{blue}{37.8}  & \cellcolor[rgb]{.851,.851,.851}\textcolor{blue}{29.8} & \cellcolor[rgb]{.851,.851,.851}30.2   & \cellcolor[rgb]{.851,.851,.851}\textcolor{blue}{40.6} & \cellcolor[rgb]{.851,.851,.851}\textcolor{blue}{44.1}  & \cellcolor[rgb]{.851,.851,.851}\textcolor{blue}{34.2}  & \cellcolor[rgb]{.851,.851,.851}29.9 & \cellcolor[rgb]{.851,.851,.851}30.5   & \cellcolor[rgb]{.851,.851,.851}\textcolor{blue}{21.3} & \cellcolor[rgb]{.851,.851,.851}\textcolor{blue}{23.0}  & \cellcolor[rgb]{.851,.851,.851}\textcolor{blue}{31.6} \\
			\cellcolor[rgb]{.851,.851,.851}STGFormer (Ours, SH, T=243)                         & \cellcolor[rgb]{.851,.851,.851}            & \cellcolor[rgb]{.851,.851,.851}\textcolor{red}{28.4} & \cellcolor[rgb]{.851,.851,.851}\textcolor{red}{32.1}  & \cellcolor[rgb]{.851,.851,.851}\textcolor{red}{28.6} & \cellcolor[rgb]{.851,.851,.851}\textcolor{red}{27.9}  & \cellcolor[rgb]{.851,.851,.851}\textcolor{red}{31.4}  & \cellcolor[rgb]{.851,.851,.851}\textcolor{red}{36.9}  & \cellcolor[rgb]{.851,.851,.851}\textcolor{red}{29.0} & \cellcolor[rgb]{.851,.851,.851}\textcolor{red}{29.3}   & \cellcolor[rgb]{.851,.851,.851}\textcolor{red}{40.0} & \cellcolor[rgb]{.851,.851,.851}\textcolor{red}{43.2}  & \cellcolor[rgb]{.851,.851,.851}\textcolor{red}{33.6}  & \cellcolor[rgb]{.851,.851,.851}\textcolor{red}{29.3} & \cellcolor[rgb]{.851,.851,.851}\textcolor{blue}{29.6}   & \cellcolor[rgb]{.851,.851,.851}\textcolor{red}{20.5} & \cellcolor[rgb]{.851,.851,.851}\textcolor{red}{22.4}  & \cellcolor[rgb]{.851,.851,.851}\textcolor{red}{30.8} \\ \hline \hline
		\end{tabular}
	}
	\label{tab:table2}
	\vspace{0mm}
\end{table*}

\begin{table*}[H]
	\centering
	\caption{Performance Comparison on Human3.6M uses ground-truth 2D poses as inputs under Protocol 1. Methods marked with * denote transformer-based approaches, \textasciicircum indicates GCN-based methods, and $\dagger$ represents graph-augmented transformer methods.}
	\vspace{0mm}
	\resizebox{0.95\textwidth}{!}{
		\begin{tabular}{cccccccccccccccccc}
			\hline \hline
			Method                                  & Publication & Dir. & Disc. & Eat. & Greet & Phone & Photo & Pose & Purch. & Sit  & SitD. & Smoke & Wait & WalkD. & Walk & WlkT. & Avg. \\ \hline
			StridedFormer \cite{li2022exploiting}* (T =243)         & TMM’22      & 27.1 & 29.4  & 26.5 & 27.1  & 28.6  & 33.0  & 30.7 & 26.8   & 38.2 & 34.7  & 29.1  & 29.8 & 26.8   & 19.1 & 19.8  & 28.5 \\
			MHFormer \cite{li2022mhformer}* (T=351)               & CVPR’22     & 27.7 & 32.1  & 29.1 & 28.9  & 30.0  & 33.9  & 33.0 & 31.2   & 37.0 & 39.3  & 30.0  & 31.0 & 29.4   & 22.2 & 23.0  & 30.5 \\
			MixSTE \cite{zhang2022mixste}* (T=243)                 & CVPR’22     & 21.6 & \textcolor{blue}{22.0}  & \textcolor{red}{20.4} & \textcolor{blue}{21.0}  & \textcolor{red}{20.8}  & \textcolor{blue}{24.3}  & 24.7 & 21.9   & \textcolor{blue}{26.9} & \textcolor{red}{24.9}  & \textcolor{red}{21.1}  & \textcolor{blue}{21.5} & 20.8   & 14.7 & 15.7  & \textcolor{blue}{21.6} \\
			P-STMO \cite{shan2022p}* (T=243)                 & ECCV’22     & 28.5 & 30.1  & 28.6 & 27.9  & 29.8  & 33.2  & 31.3 & 27.8   & 36.0 & 37.4  & 29.7  & 29.5 & 28.1   & 21.0 & 21.0  & 29.3 \\
			HSTFormer \cite{qian2023hstformer}* (T=81)               & arXiv’23    & 24.9 & 27.4  & 28.1 & 25.9  & 28.2  & 33.5  & 28.9 & 26.8   & 33.4 & 38.2  & 27.2  & 26.7 & 27.1   & 20.4 & 20.8  & 27.8 \\
			GLA-GCN \cite{yu2023gla}\textasciicircum (T=243) & CVPR’23     & 26.5 & 27.2  & 29.2 & 25.4  & 28.2  & 31.7  & 29.5 & 26.9   & 37.8 & 39.9  & 29.9  & 27.0 & 27.3   & 20.5 & 20.8  & 28.5 \\
			STCFormer \cite{tang20233d}$\dagger$ (T=243)             & CVPR’23     & \textcolor{blue}{21.4} & 22.6  & 21.0 & 21.3  & 23.8  & 26.0  & \textcolor{red}{24.2} & 20.0   & 28.9 & 28.0  & \textcolor{blue}{22.3}  & 21.4 & \textcolor{blue}{20.1}  & \textcolor{blue}{14.2} & \textcolor{red}{15.0}  & 22.0 \\
			\cellcolor[rgb]{.851,.851,.851}STGFormer (Ours, T=81)                            & \cellcolor[rgb]{.851,.851,.851}            & \cellcolor[rgb]{.851,.851,.851}22.6 & \cellcolor[rgb]{.851,.851,.851}22.3  & \cellcolor[rgb]{.851,.851,.851}21.5 & \cellcolor[rgb]{.851,.851,.851}21.2  & \cellcolor[rgb]{.851,.851,.851}23.6  & \cellcolor[rgb]{.851,.851,.851}25.3  & \cellcolor[rgb]{.851,.851,.851}24.8 & \cellcolor[rgb]{.851,.851,.851}\textcolor{blue}{19.9}   & \cellcolor[rgb]{.851,.851,.851}27.9 & \cellcolor[rgb]{.851,.851,.851}28.0  & \cellcolor[rgb]{.851,.851,.851}22.9  & \cellcolor[rgb]{.851,.851,.851}21.5 & \cellcolor[rgb]{.851,.851,.851}20.4   & \cellcolor[rgb]{.851,.851,.851}14.5 & \cellcolor[rgb]{.851,.851,.851}16.8  & \cellcolor[rgb]{.851,.851,.851}22.2 \\
			\cellcolor[rgb]{.851,.851,.851}STGFormer (Ours, T=243)                           & \cellcolor[rgb]{.851,.851,.851}            & \cellcolor[rgb]{.851,.851,.851}\textcolor{red}{21.2} & \cellcolor[rgb]{.851,.851,.851}\textcolor{red}{21.7}  & \cellcolor[rgb]{.851,.851,.851}\textcolor{blue}{20.6} & \cellcolor[rgb]{.851,.851,.851}\textcolor{red}{20.6}  & \cellcolor[rgb]{.851,.851,.851}\textcolor{blue}{22.7}  & \cellcolor[rgb]{.851,.851,.851}\textcolor{red}{24.0}  & \cellcolor[rgb]{.851,.851,.851}\textcolor{blue}{24.4} & \cellcolor[rgb]{.851,.851,.851}\textcolor{red}{19.6}   & \cellcolor[rgb]{.851,.851,.851}\textcolor{red}{26.4} & \cellcolor[rgb]{.851,.851,.851}\textcolor{blue}{27.1}  & \cellcolor[rgb]{.851,.851,.851}22.6  & \cellcolor[rgb]{.851,.851,.851}\textcolor{red}{21.3} & \cellcolor[rgb]{.851,.851,.851}\textcolor{red}{19.6}   & \cellcolor[rgb]{.851,.851,.851}\textcolor{red}{14.1} & \cellcolor[rgb]{.851,.851,.851}\textcolor{blue}{15.3}  & \cellcolor[rgb]{.851,.851,.851}\textcolor{red}{21.4} \\ \hline \hline
		\end{tabular}
	}
	\label{tab:table3}
	\vspace{0mm}
\end{table*}

\begin{table}[htbp]
	\centering
	\caption{Performance Comparison on MPI-INF-3DHP: Methods marked with * denote transformer-based approaches, \textasciicircum indicates GCN-based methods, and $\dagger$ represents graph-augmented transformer methods.}
	\vspace{0mm}
	\resizebox{0.4\textwidth}{!}{
		\begin{tabular}{ccccc}
			\hline \hline
			Method                                 & Publication & PCK$\uparrow$ & AUC$\uparrow$ & MPJPE$\downarrow$ \\ \hline
			MHFormer \cite{li2022mhformer}* (T=9)                & CVPR’22     & 93.8 & 63.3 & 58.0   \\
			MixSTE \cite{zhang2022mixste}* (T=27)                 & CVPR’22     & 94.4 & 66.5 & 54.9   \\
			P-STMO \cite{shan2022p}* (T=81)                 & ECCV’22     & 97.9 & 75.8 & 32.2   \\
			HSTFormer \cite{qian2023hstformer}* (T=81)              & arXiv’23    & 98.0 & 78.6 & 28.3   \\
			PoseFormerV2 \cite{zhao2023poseformerv2}* (T =81)          & CVPR’23     & 97.9 & 78.8 & 27.8   \\
			Einfalt et al. \cite{einfalt2023uplift}* (T=81)         & WACV’23     & 95.4 & 67.6 & 46.9   \\
			UGCN \cite{wang2020motion}\textasciicircum (T=96)    & ECCV’20     & 86.9 & 62.1 & 68.1   \\
			GLA-GCN \cite{yu2023gla}\textasciicircum (T=81) & CVPR’23     & 98.5 & 79.1 & 27.8   \\
			STCFormer \cite{tang20233d}$\dagger$ (T=81)             & CVPR’23     & \textcolor{blue}{98.7} & \textcolor{blue}{83.9} & 23.1   \\
			MotionAGFormer \cite{mehraban2024motionagformer}$\dagger$ (T=27)        & WACV’23     & 98.2 & 83.5 & \textcolor{red}{19.2}   \\
			\cellcolor[rgb]{.851,.851,.851}STGFormer (Ours, T=27)                           & \cellcolor[rgb]{.851,.851,.851}            & \cellcolor[rgb]{.851,.851,.851}98.2 & \cellcolor[rgb]{.851,.851,.851}83.8 & \cellcolor[rgb]{.851,.851,.851}\textcolor{blue}{22.6}   \\
			\cellcolor[rgb]{.851,.851,.851}STGFormer (Ours, T=81)                           & \cellcolor[rgb]{.851,.851,.851}            & \cellcolor[rgb]{.851,.851,.851}\textcolor{red}{98.8} & \cellcolor[rgb]{.851,.851,.851}\textcolor{red}{84.1} & \cellcolor[rgb]{.851,.851,.851}\textcolor{red}{19.2}   \\ \hline \hline
		\end{tabular}
	}
	\label{tab:table4}
	\vspace{0mm}
\end{table}

\noindent\textit{2) Qualitative Results:}

In this section, we conduct a qualitative analysis of the model's performance by visualizing the attention layers and assessing the effects of model on Human3.6M dataset and videos in-the-wild.

\begin{figure}[htbp]
	\centering
	\includegraphics[width=8cm]{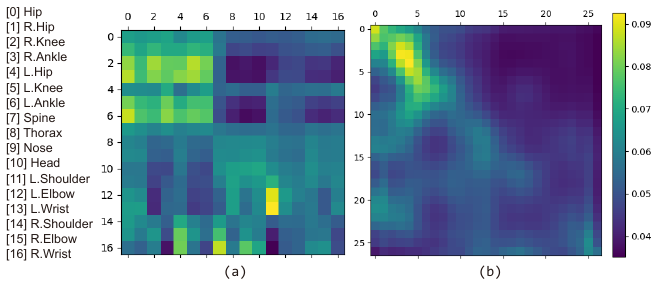}
	\caption{Visualization of attention maps from the spatial and temporal attention modules in STGFormer.}
	\vspace{-2mm}
	\label{fig4}
\end{figure}

\textbf{Attention visualization.}
The walking action from test set S11 is selected to demonstrate the attention mechanisms. The visualization of the attention layers shows the distribution of attention weights across different joints and time steps, with brighter colors indicating higher focus on the respective node during the current phase of motion. In the spatial attention map shown in Figure \ref{fig4}(a), both the horizontal and vertical axes represent joint indices. The dependencies between the left hip, knee, and ankle nodes and the right hip, knee, and ankle nodes are clearly significant for the "walking" sequence. The temporal attention layer in Figure \ref{fig4}(b) uses the x-axis and y-axis to represent different frames. This visualization reveals strong correlations between joints across adjacent frames as well as higher-order frames. Additionally, it shows that higher-order temporal connections, facilitated by self-attention, effectively capture joint temporal relationships from an alternative perspective.

\begin{figure*}[H]
	\centering
	\includegraphics[width=15cm]{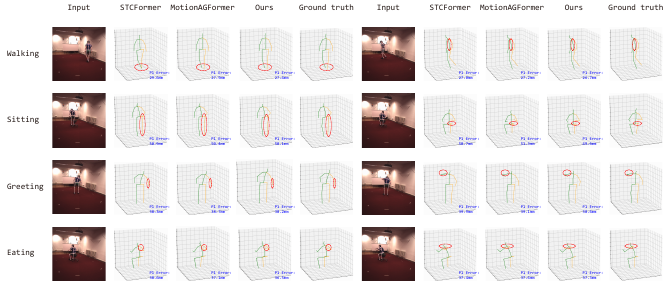}
	\caption{Examples of 3D pose estimation by our framework on Human3.6M dataset.}
	\vspace{-2mm}
	\label{fig5}
\end{figure*}

\textbf{Result visualization on Human3.6M.}
We also evaluate the visual results of the estimated poses and 3D ground truth from the Human3.6M dataset, comparing them with the state-of-the-art methods STCFormer \cite{tang20233d} and MotionAGFormer \cite{mehraban2024motionagformer} in Figure \ref{fig5}. The figure showcases four randomly selected examples from the Walking, Sitting, Greeting, and Eating sequences. Leveraging the framework's robust modeling capabilities in both temporal and spatial dimensions, our method demonstrates strong performance in handling complex actions, such as Sitting (second row) and Eating (fourth row), by effectively reconstructing poses in three-dimensional space. These results highlight that our approach yields more accurate pose estimations compared to both STCFormer and MotionAGFormer.

\begin{figure*}[H]
	\centering
	\includegraphics[width=10cm]{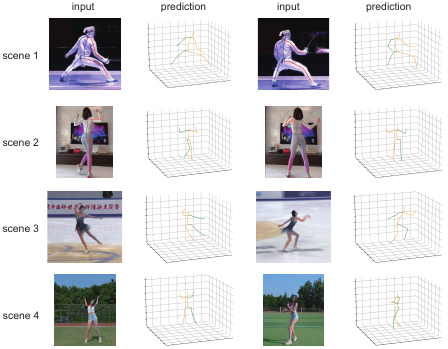}
	\caption{Examples of 3D pose estimation by our framework on in-the-wild vidoes.}
	\vspace{-2mm}
	\label{fig6}
\end{figure*}

\textbf{Result visualization on videos in-the-wild.}
Estimating 3D human poses from in-the-wild videos presents significant challenges due to complex environmental factors and unknown camera parameters. One practical way to assess a model's generalization ability is by applying a pre-trained network to such videos. In this study, we first use YOLOv3 \cite{redmon2018yolov3} to detect persons in the videos, followed by HRNet \cite{sun2019deep} for 2D keypoint detection. We then employ the pre-trained STGFormer model, trained on the Human3.6M dataset, to estimate the 3D human poses for videos sourced from Bilibili. For evaluation, we select a variety of challenging video clips, including fencing, dance, yoga, and figure skating. As shown in Figure \ref{fig6}, our method produces plausible and high-fidelity results for these in-the-wild videos, demonstrating the superior generalization capability of our approach.

\noindent\textit{3) Ablation Studies}

To further investigate the performance contributions of different components of our proposed model, we conducted a series of ablation experiments on the Human3.6M dataset, using CPN-pretrained 2D keypoint coordinates as input. The ablation studies were organized into two main parts: (1) the impact of frame count on model performance, and (2) the effect of different model components on the overall accuracy.

\textbf{Impact of frame count.}
In the first set of ablation studies, we explored how the performance of our proposed model varies with different frame counts. As illustrated in Table \ref{tab:table5}, the P1 error comparison of our model with other methods on the Human3.6M dataset is detailed for 27, 81, and 243 frames. Overall, we observed a trend of decreasing P1 error with an increasing frame count in our method and others. Among these methods, our model achieved the best results at 27, 81, and 243 frames, demonstrating strong performance in 3D pose estimation in long sequence videos. Notably, with a comparable number of parameters, our model requires less than 79\% of the FLOPs compared to the currently best-performing MixSTE \cite{zhang2022mixste} at 243 frames.

\begin{table}[htbp]
	\centering
	\caption{P1 error comparisons with different numbers of CPN-pretrained 2D keypoint frames on Human3.6M dataset. The best result in each column is highlighted in red.}
	\vspace{0mm}
	\resizebox{0.4\textwidth}{!}{
		\begin{tabular}{ccccc}
			\hline \hline
			Method                & Frames & Parameters & FLOPs (M) & P1 (mm) \\ \hline
			StridedFormer \cite{li2022exploiting} & 27     & 4.01M      & 163       & 46.9    \\
			MHFormer \cite{li2022mhformer} & 27     & 18.92M     & 1000      & 45.9    \\
			MixSTE \cite{zhang2022mixste} & 27     & 33.61M     & 15402     & 45.1    \\
			P-STMO \cite{shan2022p} & 27     & 4.6M       & 164       & 46.1    \\
			STCFormer \cite{tang20233d} & 27     & 4.75M      & 2173      & 44.1    \\
			\cellcolor[rgb]{.851,.851,.851}STGFormer                  & \cellcolor[rgb]{.851,.851,.851}27     & \cellcolor[rgb]{.851,.851,.851}12.66M     & \cellcolor[rgb]{.851,.851,.851}3104      & \cellcolor[rgb]{.851,.851,.851}\textcolor{red}{43.7}    \\ \hline
			StridedFormer \cite{li2022exploiting} & 81     & 4.06M      & 392       & 45.4    \\
			MHFormer \cite{li2022mhformer} & 81     & 19.67M     & 1561      & 44.5    \\
			MixSTE \cite{zhang2022mixste} & 81     & 33.61M     & 46208     & 42.7    \\
			P-STMO \cite{shan2022p} & 81     & 5.4M       & 493       & 44.1    \\
			STCFormer \cite{tang20233d} & 81     & 4.75M      & 6520      & 42.0    \\
			\cellcolor[rgb]{.851,.851,.851}STGFormer                  & \cellcolor[rgb]{.851,.851,.851}81     & \cellcolor[rgb]{.851,.851,.851}20.08M     & \cellcolor[rgb]{.851,.851,.851}9632      & \cellcolor[rgb]{.851,.851,.851}\textcolor{red}{41.1}    \\ \hline
			StridedFormer \cite{li2022exploiting} & 243    & 4.23M      & 1372      & 44.0    \\
			MHFormer \cite{li2022mhformer} & 243    & 24.72M     & 4812      & 43.2    \\
			MixSTE \cite{zhang2022mixste} & 243    & 33.61M     & 138623    & 40.9    \\
			P-STMO \cite{shan2022p} & 243    & 6.7M       & 1737      & 42.8    \\
			STCFormer \cite{tang20233d} & 243    & 4.75M      & 19561     & 41.0    \\
			\cellcolor[rgb]{.851,.851,.851}STGFormer                  & \cellcolor[rgb]{.851,.851,.851}243    & \cellcolor[rgb]{.851,.851,.851}31.91M     & \cellcolor[rgb]{.851,.851,.851}28871     & \cellcolor[rgb]{.851,.851,.851}\textcolor{red}{40.3}    \\ \hline \hline
		\end{tabular}
	}
	\label{tab:table5}
	\vspace{0mm}
\end{table}

\textbf{Analysis of model components.}
The second set of ablation studies examines the performance contribution of various components in our proposed model, specifically focusing on the effects of spatial and temporal graph attention, as well as different graph structures, as detailed in Table \ref{tab:table6}. In the table, STG attention (STGA) refers to the parallel combination of spatial graph attention (SGA) and temporal graph attention (TGA), which correspond to the spatial and temporal attention mechanisms in Section \ref{sec3.3}, respectively. SGA $\rightarrow$ TGA and TGA $\rightarrow$ SGA represent serial structures with reversed orders. MHR-GCN refers to the parallel dual-path structure consisting of S-GCN and T-GCN, as described in Section \ref{sec3.4}. S-GCN $\rightarrow$ T-GCN and T-GCN $\rightarrow$ S-GCN represent serial connections between the two components in different orders. GCN represents the case where no spatio-temporal splitting is performed.

Initially, when either SGA or TGA was applied independently with an MLP replacing the graph convolutional network, the P1 error increased significantly, showing values of 263.2 mm and 57.8 mm, respectively. This indicates that using attention mechanisms in isolation results in suboptimal feature representation, as both spatial and temporal dependencies need to be fully captured to improve the model's performance. When combining the spatial and temporal graph attention (STGA) in parallel, the error was reduced to 53.3 mm, demonstrating the enhanced efficacy of jointly capturing both spatial and temporal dependencies. Further analysis reveals that the order of attention mechanisms also impacts the performance. The serial configuration of SGA $\rightarrow$ TGA resulted in a P1 error of 54.1 mm, while the reverse configuration, TGA $\rightarrow$ SGA, produced a higher error of 120.5 mm. This suggests that temporal dependencies are more effectively captured before spatial ones in our framework, emphasizing the importance of modeling temporal dynamics before refining spatial relationships. The introduction of the MHR-GCN, consisting of the spatial graph convolution network (S-GCN) and temporal graph convolution network (T-GCN) in parallel, leads to a further reduction in error to 43.7 mm. This result highlights the importance of both spatial and temporal graph convolutions working together in parallel to jointly capture complex dependencies, thereby providing more accurate feature representations. Interestingly, when combining S-GCN and T-GCN in serial (either S-GCN $\rightarrow$ T-GCN or T-GCN $\rightarrow$ S-GCN), the errors were slightly higher (52.5 mm and 47.0 mm, respectively), which suggests that parallel pathways are more effective than serial ones for processing spatial and temporal information simultaneously. These results indicate that each component plays a crucial role in improving model performance. The spatial and temporal graph attention mechanisms are essential for capturing the respective dependencies, while the parallel S-GCN and T-GCN structure offers a more robust representation of human pose dynamics compared to serial connections.

\begin{table}[htbp]
	\centering
	\caption{Performance contribution of each component in the proposed framework on Human3.6M dataset.}
	\vspace{0mm}
	\resizebox{0.3\textwidth}{!}{
		\begin{tabular}{ccc}
			\hline\hline
			attention part                             & graph part                                    & P1 (mm) \\ \hline
			\multicolumn{1}{l}{SGA}                    & \multicolumn{1}{c}{\multirow{5}{*}{MLP}}      & 263.2   \\
			\multicolumn{1}{l}{TGA}                    & \multicolumn{1}{c}{}                          & 57.8    \\ 
			\multicolumn{1}{l}{SGA $\rightarrow$ TGA}  & \multicolumn{1}{c}{}                          & 54.1    \\ 
			\multicolumn{1}{l}{TGA $\rightarrow$ SGA}  & \multicolumn{1}{c}{}                          & 120.5   \\ 
			\multicolumn{1}{l}{STGA}                   & \multicolumn{1}{c}{}                          & 53.3    \\ 
			\hline
			\multicolumn{1}{l}{\multirow{6}{*}{STGA}}  & \multicolumn{1}{l}{GCN}                       & 50.6    \\ 
			\multicolumn{1}{l}{}                       & \multicolumn{1}{l}{S-GCN}                     & 46.9    \\ 
			\multicolumn{1}{l}{}                       & \multicolumn{1}{l}{T-GCN}                     & 45.0    \\ 
			\multicolumn{1}{l}{}                       & \multicolumn{1}{l}{S-GCN $\rightarrow$ T-GCN} & 52.5    \\  
			\multicolumn{1}{l}{}                       & \multicolumn{1}{l}{T-GCN $\rightarrow$ S-GCN} & 47.0    \\  
			\multicolumn{1}{l}{}                       & \multicolumn{1}{l}{MHR-GCN}                   & 43.7    \\ 
			\hline\hline
		\end{tabular}
	}
	\label{tab:table6}
	\vspace{0mm}
\end{table}

\section{Conclusion}

In this paper, we introduce Spatio-Temporal GraphFormer, a novel model that integrates long-range temporal-spatial dependencies with graph structural information in its attention mechanisms. It utilizes a dual-path modulated hop-wise regular GCN module to enhance graph information extraction for 3D human pose estimation in videos. Specifically, the spatio-temporal criss-cross graph attention separately models spatial and temporal correlations, leveraging encoded graph structure information. The dual-path modulated hop-wise regular GCN module adjusts graph weights and accounts for higher-order dependencies across both time and space dimensions. Experiments on two benchmark datasets affirm the efficacy and superior generalization capability of our proposed model compared to state-of-the-art techniques. While the proposed method presents several promising innovations for improving 3D human pose estimation, it still has limitations in multi-person and occlusion scenarios. First, the model assumes a single-person context, which could lead to performance degradation in frames with multiple individuals. The attention mechanism may struggle to effectively distinguish overlapping poses, leading to confusion in keypoint estimation. Additionally, occlusions, where parts of the body are hidden, present a significant challenge. Although spatiotemporal graph attention captures long-range dependencies, occlusions can disrupt the model’s ability to accurately infer 3D pose information, especially when joints are partially visible. Thus, while the method performs well in controlled, single-person scenarios, its robustness in crowded or occluded environments requires further exploration and enhancement. Addressing these limitations would improve the model's applicability in diverse and practical settings. For future work, we aim to develop a multi-person skeletal sequence spatio-temporal graph representation method, extending our model’s applicability to 3D human pose estimation in multi-person scenarios and broadening its use across various contexts.

\clearpage










\bibliographystyle{elsarticle-num}

\bibliography{refs}



\end{document}